\documentclass{article}
\usepackage{ijcai25}

\usepackage{times}
\usepackage{soul}
\usepackage{url}
\usepackage[hidelinks]{hyperref}
\usepackage[utf8]{inputenc}
\usepackage[small]{caption}
\usepackage{graphicx}
\usepackage{amsmath}
\usepackage{amsthm}
\usepackage{booktabs}
\usepackage[switch]{lineno}

\usepackage{amsxtra, amsfonts, amssymb, amstext, mathtools}
\usepackage{nicefrac}
\usepackage{xspace}
\usepackage{graphics,color}
\usepackage{wrapfig}
\usepackage{subcaption}
\usepackage{epstopdf}
\usepackage[algo2e,ruled,vlined,linesnumbered]{algorithm2e}\SetArgSty{upshape}

\newtheorem{theorem}{Theorem}
\newtheorem{lemma}[theorem]{Lemma}

\newtheorem{definition}[theorem]{Definition}
\DeclareMathOperator{\age}{age}

\newcommand{\NSGA}{\mbox{NSGA}\nobreakdash-II\xspace}
\newcommand{\NSGAT}{\mbox{NSGA}\nobreakdash-III\xspace}
\newcommand{\SMS}{\mbox{SMS-EMOA}\xspace}

\newcommand{\jump}{\textsc{Jump}\xspace}
\newcommand{\oneminmax}{\textsc{OneMinMax}\xspace}

\newcommand{\cocz}{\textsc{COCZ}\xspace}

\newcommand{\lotz}{\textsc{LOTZ}\xspace}
\newcommand{\ojzj}{\textsc{OJZJ}\xspace}
\newcommand{\mojzj}{m\textsc{OJZJ}\xspace}

\newcommand{\R}{\ensuremath{\mathbb{R}}}

\newcommand{\N}{\ensuremath{\mathbb{N}}} 
\newcommand{\Z}{\ensuremath{\mathbb{Z}}}

\let\originalleft\left
\let\originalright\right
\renewcommand{\left}{\mathopen{}\mathclose\bgroup\originalleft}
\renewcommand{\right}{\aftergroup\egroup\originalright}

\title{Scalable Speed-ups for the \SMS from a Simple Aging Strategy}

\author{
Submission to IJCAI 2025, Paper ID 6901}

\author{
Mingfeng Li$^1$
\and
Weijie Zheng$^1$\footnote{Corresponding author.}\and
Benjamin Doerr$^{2}$\\
\affiliations
$^1$ School of Computer Science and Technology, National Key Laboratory of Smart Farm Technologies and Systems, International Research Institute for Artificial Intelligence, 
\\
Harbin Institute of Technology, Shenzhen\\
$^2$Laboratoire d'Informatique (LIX), CNRS, \'Ecole Polytechnique, \\
Institut Polytechnique de Paris, Palaiseau, France\\
\emails
\{dengrenzhong, zhengweijie\}@hit.edu.cn
}

\begin{document}

\maketitle
\begin{abstract}
Different from single-objective evolutionary algorithms, where non-elitism is an established concept, multi-objective evolutionary algorithms almost always select the next population in a greedy fashion. In the only notable exception, Bian, Zhou, Li, and Qian (IJCAI 2023) proposed a stochastic selection mechanism for the SMS-EMOA and proved that it can speed up computing the Pareto front of the bi-objective jump benchmark with problem size $n$ and gap parameter $k$ by a factor of $\max\{1,2^{k/4}/n\}$. While this constitutes the first proven speed-up from non-elitist selection, suggesting a very interesting research direction, it has to be noted that a true speed-up only occurs for $k \ge 4\log_2(n)$, where the runtime is super-polynomial, and that the advantage reduces for larger numbers of objectives as shown in a later work. 
In this work, we propose a different non-elitist selection mechanism based on aging, which exempts individuals younger than a certain age from a possible removal. 
This remedies the two shortcomings of  stochastic selection: We prove a speed-up by a factor of $\max\{1,\Theta(k)^{k-1}\}$, regardless of the number of objectives.
In particular,  a  positive speed-up can already be observed for constant $k$, the only setting for which polynomial runtimes can be witnessed.
Overall, this result supports the use of non-elitist selection schemes, but suggests that aging-based mechanisms can be considerably more powerful than stochastic selection mechanisms.
\end{abstract}

\section{Introduction}
In many real-world optimization problems, several objectives have to be optimized simultaneously. 
For such multi-objective optimization problems, evolutionary algorithms have been employed with great success~\cite{ZhouQLZSZ11}, e.g., the famous \NSGA algorithm~\cite{DebPAM02} has received more than 50,000 citations on Google scholar.

Interestingly, different from single-objective optimization, these multi-objective evolutionary algorithms (MOEAs) almost always select the next population in a greedy fashion. The \NSGA, for example, uses non-dominated sorting to select the first fronts into the next population. As tie-breaker for the critical front, the crowding distance is used, and again the individuals with largest crowding distance survive. Similarly, the \NSGAT \cite{DebJ14} and \SMS \cite{BeumeNE07} greedily select according to non-dominated sorting, using references points or the hypervolume contribution as the secondary (greedy) selection criteria.

This greedy behavior in multi-objective evolutionary computation differs significantly from single-objective evolutionary computation, where concepts such as \emph{selection pressure} are used to quantify the degree of greediness of the selection. Furthermore, classic selection operators such as tournament selection or roulette-wheel selection, are intentionally designed to be less greedy than truncation selection.

The first to substantially challenge the believe in greedy selection in multi-objective evolutionary computation were Bian et al.~\shortcite{BianZLQ23}, who recently proposed a stochastic population update for the \SMS. Instead performing a greedy selection over the whole combined parent and offspring population, they let (roughly) a random half of the individuals survive irrespective of their quality and conduct the usual selection of the \SMS only in the other half. With this non-greedy selection mechanism, they obtained a proven speed-up by a factor of $\max\{1, 2^{k/4}/n\}$ for the bi-objective \ojzj benchmark with problem size $n$ and gap parameter~$k$ (this speed-up was later improved to $\max\{1,2^k/n\}$ in~\cite{ZhengD24}).\footnote{Another non-elitist attempt was studied by Zheng et al.~\shortcite{ZhengLDD24}. Inspired by the theoretical advantage of the Metropolis algorithm for the single-objective DLB benchmark~\cite{WangZD24}, they discussed how to use the Metropolis algorithm for multiple objectives. While that work could detect strengths and weaknesses of different variants of the multi-objective Metropolis algorithm,  compared to other MOEAs none of these variants could achieve better theoretical runtime guarantees for the bi-objective DLB benchmark.}

Following the exciting results on the stochastic population update of Bian et al.~\shortcite{BianZLQ23}, Zheng and Doerr~\shortcite{ZhengD24} (among other results) discussed how these results extend to more than two objectives. For the $m$-objective \mojzj benchmark, they proved a speed-up factor of (roughly) $\max\{1,2^k/n^{m/2}\}$. Hence the proven speed-up from the stochastic population update vanishes for larger numbers of objectives.

We note here that from the practical perspective,~\cite{BianZLQ23} might have two drawbacks: working on artificial benchmarks and highly expensive computing time inherited from the usage of hypervolume. We agree that it is more interesting to analyze practical problems. However, as is widely known, it is a bitter trade-off among the practical usage and the success possibility of establishing theory for the analyzed problem. Besides, the \jump series benchmark analyzed in~\cite{BianZLQ23} is a widely used multimodal class in theory community, which inspires many interesting findings, such as the efficiency of the crossover~\cite{DoerrQ23crossover}, the advantage of the estimation of distribution algorithms~\cite{Witt23}. Moreover, it is questionable to generalize an algorithm to more practical problems if it cannot efficiently solve such widely analyzed toy problems in the theory community. Hence, before we consider the practical problems, we in this theoretical paper, still stick to this \jump benchmark series in two- and more- objective settings.

In terms of the second drawback of highly expensive computing time, we admit the time-consuming procedure of computing the hypervolume in the algorithm (including the \SMS and its variant)~\cite{guerreiro2021hypervolume}. Note that the actual computing time depends on the execution time in each iteration (which relies on the detailed implementation of the hypervolume), as well as the number of iterations to reach a goal (which is the runtime complexity analyzed in theoretical computer science). Both are ways for the efficient applications of the algorithm but on different aspects. Following~\cite{BianZLQ23} and other papers in the theory community, we focus on the theoretical runtime complexity perspective to address the open question whether a non-elitist survival strategy can result in scalable speed-ups for all number of objectives. We will not touch how to better implement the calculation of the hypervolume in this work.

\textbf{Our contributions:} In this paper, we take such an attempt and design a simple aging strategy for the population update process of the \SMS. We initialize each individual with the old enough age $\tau$. Any newly generated individual has an age of $0$. Only individuals with an age of at least $\tau$ will be subject to the original survival selection of the \SMS. After the removal, all remaining individuals will increase their ages by $1$ and enter into the next generation. This simple aging strategy allows any newly generated individual, irrespective of its quality, to survive for at least $\tau$ iterations. 

For the \SMS with this selection strategy, we prove that the runtime for covering the full Pareto front of the bi-objective \ojzj benchmark is $O(n^{k+1} / \Theta(k)^{k-1})$, see Theorem~\ref{thm:ojzj}, comparing favorably with the best known guarantee of $O(n^{k+1})$ for the classic \SMS~\cite{BianZLQ23}. 
For the $m$-objective version \mojzj, we show a guarantee of $O(\overline{M}km n^k / \Theta(k)^k)$, see Theorem~\ref{thm:mojzj}, and note that the best known upper bound for the classic \SMS is $O(\overline{M}mn^k)$ by Wietheger and Doerr~\cite{WiethegerD24}, where $\overline M$ denotes the size of the largest set of pairwise non-dominating solutions of the problem. Hence, a speed-up factor of $\max\{1,\Theta(k)^{k-1}\}$ 
has been obtained for the $m$-objective \mojzj benchmark. 

Our experimental results further support these theoretical findings, demonstrating that even for small values of $k$ in both $\ojzj$ and $\mojzj$, the aging strategy significantly accelerates the \SMS algorithm compared to both the original version and the variant with the stochastic population update.

\section{Preliminaries}
\subsection{Multi-Objective Optimization}
In multi-objective optimization, the goal is to find a set of optimal trade-off solutions that effectively balance multiple conflicting objectives, typically represented as a Pareto set. As common in the theory community, this paper considers the multi-objective pseudo-Boolean maximization problems, that is, to maximize \( f(x) = (f_1(x), f_2(x), \dots, f_m(x)) \) subject to \( x \in \{0,1\}^n \) where $m \in\N$ represents the number of objectives, and $n\in\N$ denotes the problem size. A key concept in multi-objective optimization is Pareto dominance as not all solutions are comparable. 
We say that a solution $x_1 \in \{0,1\}^n$ \emph{weakly dominates} another solution $x_2 \in \{0,1\}^n$ (denoted as $x_1 \succeq x_2$) if $f_i(x_1) \geq f_i(x_2)$ for all $i \in \{1,\ldots,m\}$. In this case, $x_1$ is at least as good as $x_2$ in all objectives. If, in addition, $x_1$ is strictly better than $x_2$ in at least one objective (i.e., $f_i(x_1) > f_i(x_2)$ for some $i$), then we call $x_1$ \emph{dominates} $x_2$ (denoted as $x_1 \succ x_2$). A solution $x$ is called \emph{Pareto optimal} if it is not dominated by any other solutions in $\{0,1\}^n$. The set of all Pareto optimal solutions is the \emph{Pareto set}, and the corresponding function values form the \emph{Pareto front}. 

In the theory of MOEAs, the \emph{runtime} is typically defined as the number of function evaluations required for the algorithm's population to fully cover the Pareto front~\cite{AugerD11,ZhouYQ19,DoerrN20}. 
Additionally, in this paper, we will use $|x|_1$ to denote the number of ones in $x$, $|x|_0$ to denote the number of zeros in $x$, and $[a..b]$ to represent the set $\{a,a+1,...,b\}$ for $a\leq b$ and $a,b \in \Z$. We also use $\overline{M}$ to denote the maximum size of a set of incomparable solutions for a given optimization problem.

\subsection{The \SMS and Stochastic Population Update}
The \SMS is a steady-state variant of the \NSGA, which replaces the crowding distance by the hypervolume contribution as the secondary selection criterion. This algorithm operates with a fixed population size $\mu$. In each iteration, a single offspring is generated and added to the combined parent and offspring population $R_t$.  To maintain the fixed population size, one individual from $R_t$ will be removed. It first uses the non-dominated sorting procedure to partition $R_t$ into several fronts $F_1, \dots, F_{i^*}$, where $F_i$ contains all non-dominated individuals in $R_t\setminus \cup_{j=1}^{i-1}F_j$. Within the critical front $F_{i^*}$, the individual that contributes the least to the hypervolume will be removed (broken tie randomly). The hypervolume of a set $S$ of individuals w.r.t. reference point $r$ is calculated as $\text{HV}_r(S) = \mathcal{L}\left(\bigcup_{u \in S} \left\{ h \in \mathbb{R}^m \mid r \leq h \leq f(u) \right\}\right)$ where $\mathcal{L}$ represents the Lebesgue measure. The hypervolume contribution of an individual $x \in F_{i^*}$ is defined by $\Delta_r(x, F_i^*) := \text{HV}_r(F_i^*) - \text{HV}_r(F_i^* \setminus \{x\})$. 

See Algorithm~\ref{alg:sms} for the whole procedure of the classic \SMS.
As discussed before, this work will not focus on how to efficiently implement the calculation of the hypervolume, but consider how to speed-up the runtime complexity (the number of iterations to cover the full Pareto front) with the non-elitist strategy.

\begin{algorithm2e}[tb]
\caption{\SMS}
 Generate $P_0$ by selecting $\mu$ solutions uniformly and randomly from $\{0, 1\}^n$ with replacement\;
\label{alg:sms}
\For {$t=0,1,2,\dots,$}{
 Select a solution $x$ uniformly at random from $P_t$\;
 Generate $x'$ by flipping each bit of $x$ independently with probability $1/n$\;
 Use fast-non-dominated-sort()~\cite{DebPAM02} to divide $R_t=P_t\cup \{x'\}$ into $F_1,\dots,F_{i^*}$ \label{stp:fronts}\;
 Calculate $\Delta_r(z,F_{i^*})$ for all $z\in F_{i^*}$ and find $D=\arg\min_{z\in F_{i^*}}{\Delta_r(z,F_{i^*})}$\;
 Uniformly at random pick $z'\in D$ and $P_{t+1}=R_t\setminus \{z'\}$ \label{stp:rm}\;
}
\end{algorithm2e}

The stochastic population update of the \SMS, proposed in \cite{BianZLQ23}, only has one change compared to the classic \SMS. It uniformly at random selects half of the combined population for the survival selection, and the other half directly enters into the next generation. With this strategy, the inferior solutions have the chance (like not being chosen to the survival selection) to the next generation. For a clear comparison with our proposed strategy, we also state its procedure in Algorithm~\ref{alg:sms-spu}.

\begin{algorithm2e}[tb]
\caption{\SMS with the Stochastic Population Update}
\label{alg:sms-spu}
 Generate $P_0$ by selecting $\mu$ solutions uniformly and randomly from $\{0, 1\}^n$ with replacement\;
\For {$t=0,1,2,\dots,$}{
 Select a solution $x$ uniformly at random from $P_t$\;
 Generate $x'$ by flipping each bit of $x$ independently with probability $1/n$\;
 $R_t = \lfloor P_t\cup \{x'\}/2 \rfloor$ solutions uniformly and randomly selected from $P_t\cup \{x'\}$ without replacement\;
 Use fast-non-dominated-sort()~\cite{DebPAM02} to divide $R_t$ into $F_1,\dots,F_{i^*}$\;
 Calculate $\Delta_r(z,F_{i^*})$ for all $z\in F_{i^*}$ and find $D=\arg\min_{z\in F_{i^*}}{\Delta_r(z,F_{i^*})}$\;
 Uniformly at random pick $z'\in D$ and $P_{t+1}=P_t\cup \{x'\}\setminus \{z'\}$\;
}
\end{algorithm2e}

As introduced before, we note here that a speed-up by a factor of $\max\{1,2^{k/4}/n\}$ is proven in~\cite{BianZLQ23} (which is later proven to be $\max\{1,2^{k}/n\}$ in~\cite{ZhengD24}) for the bi-objective \ojzj benchmark. We also note here, as introduced before as well, that a speed-up factor of only $\Theta(1)$ for large $m$ (say $m>k$) is proven for the $m$-objective \mojzj benchmark~\cite{ZhengD24}. As discussed before, before considering the wide applications beyond the \mojzj benchmark class, we in this paper need to tackle the challenge that the existing non-elitist strategy cannot result in a scalable speed-up when the number of objectives increases, even for this widely analyzed \jump benchmark series.

\section{The \SMS with the Aging Strategy}\label{sec:aging}

As mentioned before, the non-elitist stochastic population update cannot scale well from two objectives to more objectives. This section will introduce our non-elitist aging strategy and employ it into the \SMS for the easy comparison against the stochastic population update~\cite{BianZLQ23} in the later sections. 

\subsection{Aging Strategy}
Aging has been used in different kinds of randomized search heuristics, such as in evolutionary algorithms and artificial immune systems, as noted in ~\cite{horoba2009maximal,jansen2011role,CorusOY20}. Aging operators assign each solution in the population an individual age, which increases by 1 with each generation.  Typically a maximum lifespan $\tau$ is defined, and any solution exceeding this age is removed from the population~\cite{zarges2020}. We resort to this strategy but with some modifications. Instead of the immediate removal, only individuals with the age at least $\tau$ will join the survival selection and have the chance to be removed. This is the key simple aging strategy we use to let the inferior solutions have the chance to enter into the next generations, making the algorithm not shortsighted.

In this work, we choose the \SMS to equip with this strategy, since the efficiency of the \SMS with the non-elitist stochastic population update is proven for bi-objective \ojzj~\cite{BianZLQ23}. In detail, we first initialize $\mu$ individuals randomly, and assign all individuals with the old enough age $\tau$. This assignment of age does not impact the performance of this algorithm.
In each generation, one offspring individual is generated and assigned with the age of $0$. Among the $\mu+1$ combined parent and offspring individuals, the individuals with the age at least $\tau$ will be chosen to form the multiset $R_t$, to which the original survival selection will be applied. After the survival selection, the ages of all remaining individuals in the population are incremented by 1. All other procedures in the algorithm remain consistent with the original \SMS. The details are shown in Algorithm~\ref{alg:sms-emoa-Aging}.
Obviously, this strategy ensures that any newly generated individual, including the inferiors, can survive for at least $\tau$ generations.

\begin{algorithm2e}[tb]
\caption{\SMS with the Aging Strategy}
\label{alg:sms-emoa-Aging}
Initialize $P_0$ with $\mu$ individuals chosen independently and uniformly at random from $\{0,1\}^n$ with replacement and set their age to $\tau$\;
\For{$t = 0, 1, 2, \ldots,$}{
Select $x \in P_{t}$ uniformly at random\;
Generate $y$ from $x$ via standard bit-wise mutation and set its age $y.age =0$\;
$R_t = \{ x \mid x \in  P_t \cup \{y\} \text{ and } x.\age  \geq \tau\}$\label{stp:rt}\;
Use fast-non-dominated-sort() to divide $R_t$ into $F_1,\dots,F_{i^*}$\;
Let $D=\arg \min_{z\in F_i^*} \Delta_r(z,F_{i^*})$\;
Uniformly at random pick $z' \in D$ \;
$P_{t+1} = P_t \cup \{y\} \setminus \{z'\}$\;
Raise the ages of solutions in $P_{t+1}$ by 1\;
}
\end{algorithm2e}

\subsection{Basic Behavior}
The following lemma gives the basic behavior of the \SMS with the aging strategy. That is, if the population size is at least the same as the size of largest set of incomparable solutions plus $\tau$ and 1, that is, $\overline{M} + 1 + \tau$, then any individual in the current population will have future individuals that weakly dominate it.
Due to the space limitation, we put all proofs into the supplementary material that is also submitted to the IJCAI system.

\begin{lemma}
    Consider any $m$-objective optimization problem. Consider using the \SMS with the aging strategy and with population size $\mu \geq \overline{M} + 1 + \tau$ to solve this problem. If $P_t$ contains a solution $x$, then at any later time $t'>t$, the population $P_{t'}$ will contain a solution $y$ such that $y \succeq x$. In particular, any Pareto front point once reached will be maintained in all future generations.
    \label{lem:survival}
\end{lemma}

This important behavior of not being worse will be frequently used for the runtime analysis for the Pareto front coverage.

\section{Runtime for Bi-Objective \ojzj}
Recall that the stochastic population update shows a speed-up by a factor of $\max \{1,2^{\Theta(k)}/n\}$  for the \SMS optimizing the bi-objective \ojzj benchmark~\cite{BianZLQ23}. In this section, we will also show that our proposed aging strategy can speed up the \SMS by a factor of $\max \{1,\Theta(k)^{k-1}\}$, which is also faster than the one with the stochastic population update by a factor of $\max\{1,\Theta(k)^kn/k\}$.

\subsection{\ojzj}
\jump functions \cite{DrosteJW02} are the most extensively studied multimodal benchmark in the theory of randomized search heuristics, particularly in evolutionary algorithms~\cite{BamburyBD24}. They represent a class of problems with a tunable gap parameter $k$, which reflects different depths from the local optima to the global optimum.
The \ojzj benchmark proposed by Doerr and Zheng~\cite{DoerrZ21aaai} is a bi-objective counterpart of the \jump functions with problem size $n$ and jump size $k$, which has been widely used to understand the theoretical behavior of the MOEAs for multimodal problems~\cite{DoerrQ23tec,DoerrQ23LB,DoerrQ23crossover,BianZLQ23,DoerrIK25}.
The \ojzj benchmark involves two objectives, that is, one is the \jump benchmark, and the other is the \jump function applied to $\bar{x}=1-x$. See the formal definition in the following.
\begin{definition}[\cite{DoerrZ21aaai}]
Let $n \in \mathbb{N}$ and $k=[1..n]$. The function $\ojzj_{n,k} = (f_1, f_2) : \{0, 1\}^n \to \mathbb{R}^2$ is defined by
\[
f_1(x) =
\begin{cases}
k + |x|_1, & \text{if } |x|_1 \leq n - k \text{ or } x = 1^n, \\
n - |x|_1, & \text{else};
\end{cases}
\]
\[
f_2(x) =
\begin{cases}
k + |x|_0, & \text{if } |x|_0 \leq n - k \text{ or } x = 0^n, \\
n - |x|_0, & \text{else}.
\end{cases}
\]    
\label{def:ojzj}
\end{definition}

\begin{figure}[t]
    \centering
    \includegraphics[width=1.0\columnwidth]{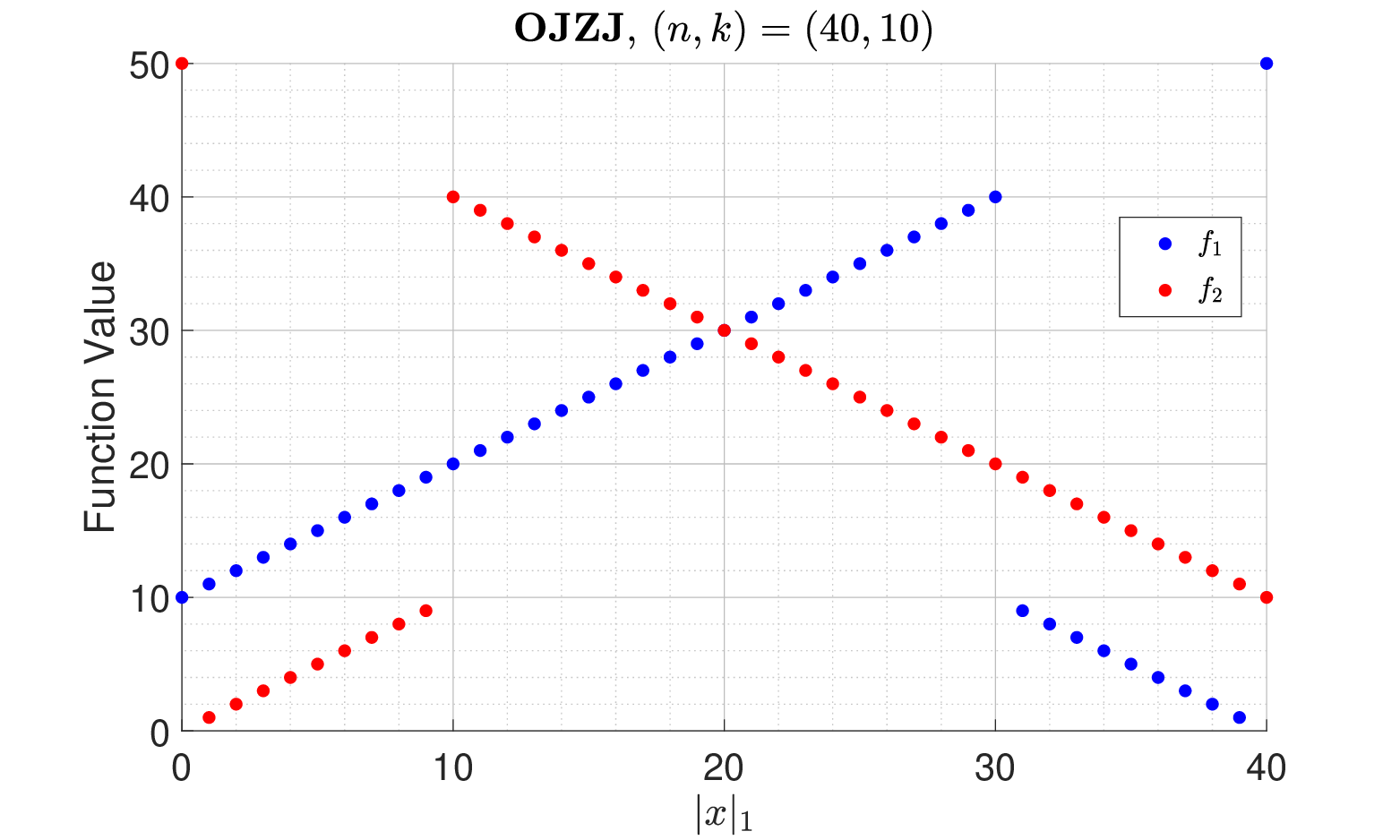}
    \caption{\ojzj with $\{n,k\} = \{40,10\}$.}
    \label{fig:f_OJZJ}
\end{figure}

Figure~\ref{fig:f_OJZJ} illustrates the characteristics of the \ojzj benchmark. As proven in \cite{DoerrZ21aaai}, the Pareto set is $\{x \mid |x|_1 \in [k..n - k] \cup \{0, n\}\}$ and the Pareto front is $\{(a, 2k + n - a) \mid a \in [2k..n] \cup \{k, n + k\}\}$. If $k \leq n/2$, the size of the largest set of incomparable individuals,  $\overline{M}$, is $n-2k+3$. Since the objective values of \ojzj are all greater than 0, we set the reference point for the \SMS to $(-1,-1)$. As mentioned before, we will calculate the runtime to cover the full Pareto front. 
 
\subsection{Runtime}
The optimization process can be divided into two phases. The first phase ends when all inner Pareto front points $\{(a,2k+n-a)\mid a\in[2k..n]\}$ are covered. Then the second phase starts, and ends when the full Pareto front is covered for the first time. Lemma~\ref{lem:OJZJ inner part} considers the runtime of the first phase, showing that after $O\left(\mu n\log n\right)+k\tau \left(\frac{2e}{1-\exp\left(-\frac{\tau}{\mu}\right)}\right)^k$ iterations in expectation, the population will cover the inner Pareto front. The proof idea is that once the population includes at least one solution in the inner Pareto optimal set $\{x\mid |x|_1\in[k..n-k]\}$, then its Hamming neighbor belonging to the inner Pareto optima can be generated by flipping one of the 0-bits or one of the 1-bits. According to Lemma~\ref{lem:survival}, any reached Pareto front point will be remained. By repeating this process and selecting the appropriate individuals, the entire inner Pareto front can be covered. 
\begin{lemma}
    Let $k\leq n/2$. Consider using the \SMS with the aging strategy and with population size $\mu \geq \overline{M} +1+ \tau$ to optimize the $\ojzj$ problem, then the expected number of iterations is 
    at most $O\left(\mu n\log n\right)+k\tau \left(\frac{2e}{1-\exp\left(-\frac{\tau}{\mu}\right)}\right)^k$
    to cover all inner Pareto front points.
    \label{lem:OJZJ inner part}
\end{lemma}

The following lemma provides the runtime analysis of the algorithm for the second phase which starts with a population covering the inner Pareto front and ends when the full Pareto front is covered. In other words, we need to find the two extreme solutions $1^n$ and $0^n$ during the phase. The proof uses a waiting time argument. That is, we first calculate the probability $p$ to find the desired $1^n$ (or $0^n$) from the boundary of the inner Pareto front within at most $k\tau$ iterations, and subsequently bound the expected waiting time by $\frac{k\tau}{p}$.
We divide $k\tau$ iterations into $k$ steps and within each step of at most $\tau$ generation, we calculate the probability of an improvement. Note that the aging strategy ensures that any newly generated solution survives for at least $\tau$ generations. Then we easily obtain the overall probability $p$ via multiplying the probabilities in all steps for their independence, and thus the runtime argument is obtained.

\begin{lemma}
    Let $k\leq n/2$. Consider using the \SMS with the aging strategy and with population size $\mu \geq \overline{M} +1+\tau$ to optimize $\ojzj$. Assume that the current population covers all inner Pareto front points. Then the full Pareto front will be covered in at most $k\tau\left(\frac{e^2 n}{\left(1-\exp\left(-\frac{\tau}{\mu}\right)\right)k}\right)^k$ iterations in expectation.   
    \label{lem:OJZJ second phase}
\end{lemma}

Combining Lemmas~\ref{lem:OJZJ inner part} and ~\ref{lem:OJZJ second phase}, we derive the runtime of the \SMS with the aging strategy stated in the following theorem.

\begin{theorem}\label{thm:ojzj}
    Let $k\leq n/2$. Consider using the \SMS with the aging strategy and with population size $\mu \geq \overline{M} +1+\tau$ to optimize $\ojzj$, then the expected iterations to cover the full Pareto front is at most 
    \begin{align*}
        &O\left(\mu n\log n\right) + \frac{k\tau\left(\left(2e\right)^k+\left(\frac{e^2n}{k}\right)^k\right)}{\left(1-\exp\left(-\frac{\tau}{\mu}\right)\right)^k} \\
        & = O\left(k\tau\left(\frac{e^2 n}{\left(1-\exp\left(-\frac{\tau}{\mu}\right)\right)k}\right)^k\right).
    \end{align*}
\end{theorem}

Note that the expected runtime of the original \SMS is $O(n^{k+1})$ for $\mu =\Theta(n)$ ~\cite{BianZLQ23}.
We set $\tau=\Theta(n)$ and $\mu = \Theta(n-2k+4+\tau)$ to also ensure $\mu = \Theta(n)$. In this setting, from Theorem~\ref{thm:ojzj}, we know that the expected iterations for the \SMS with the aging strategy to cover the full Pareto front is $O(n^{k+1}/( \Theta(k))^{k-1})$. Hence, we see a speed-up by a factor of $\max\{1,\Theta(k)^{k-1}\}$. Also noting $O(n^{k+2}/2^{\Theta(k)})$ expected number of iterations for the stochastic population update~\cite{BianZLQ23,ZhengD24}, we see our aging strategy surpasses this by a factor of $\max\{1,\Theta(k)^kn/k\}$.

\section{Runtime for $m$-Objective \mojzj}

In the previous section, we demonstrated that the \SMS with the aging strategy outperforms the original \SMS and the one with the stochastic population update on the bi-objective \ojzj problem. As pointed out before,~\cite{ZhengD24} proved that the speed-up of the stochastic population update mechanism cannot scale well from two objectives to more objectives. This section will show that our simple aging strategy  performs well for many objectives, and achieves the speed-up by a factor of $\max\{1,\Theta(k)^{k-1}\}$ against the original \SMS as well as the one with the stochastic population update. 

\subsection{\mojzj}
The \mojzj benchmark proposed by~\cite{ZhengD24} is a $m$-objective counterpart of the bi-objective \ojzj benchmark~\cite{DoerrZ21aaai}, and is the first multimodal many-objective benchmark proposed for theoretical analysis. Zheng and Doerr~\cite{ZhengD24} proved that SMS-EMOA can compute the full Pareto front of this benchmark in an expected number of $O(\mu Mn^k)$ iterations, where $M$ is the size of the Pareto front. Later, Wietheger and Doerr ~\cite{WiethegerD24} proved near-tight runtime guarantees for the SEMO, GSEMO and \SMS algorithms on this multimodal benchmark class and other popular unimodal many-objective variants of \oneminmax, \cocz, and \lotz. 
The \mojzj benchmark involves $m$ objectives. The bit string of length $n$ is divided into $m/2$ blocks, each of length $2n/m$. For each block, a bi-objective \ojzj problem is defined. See the following definition. 
\begin{definition}[\cite{ZhengD24}]
Let $m$ be the even number of objectives. Let the problem size $n$ be
a multiple of $m/2$. Let $n' = \frac{2n}{m} \in \mathbb{N}$ and $k \in [1..n']$. Let $B_i:=[(i-1)n'+1..in']$ for $i\in[1..m/2]$ denote the $i$-th block of the $n$ bit positions. Then for any $x = (x_1, \ldots, x_n)$, the $\mojzj_k=\left(f_1(x),\ldots,f_m(x)\right): \{0,1\}^n\rightarrow \R^m$ is defined by 
\[
\left(f_{2i}, f_{2i-1}(x)\right)=\ojzj(x_{B_i}), i\in[1..m/2],
\]
where the \ojzj function is defined in Definition~\ref{def:ojzj}.
\end{definition}

The Pareto set of \mojzj is $\{x \in \{0, 1\}^n \mid \forall i \in [1..m/2], \, |x_{B_i}|_1 \in [k..n' - k] \cup \{0, n'\}\}$ and the Pareto front is $\{(a_1, n' + 2k - a_1, \ldots, a_{m/2}, n' + 2k - a_{m/2}) \mid a_1, \ldots, a_{m/2} \in [2k..n'] \cup \{k, n' + k\}\}$. Hence, the size of the Pareto front is $\left(n'-2k+3\right)^{\frac{m}{2}}$. Different from $\ojzj$, the size of the largest set of incomparable solutions is not equal to the size of the Pareto front, and a trivial upper bound of $(n'+1)^\frac{m}{2}$ is given in~\cite{ZhengD24arxiv} when $k\leq n'/2$, $\overline{M}\leq(n'+1)^\frac{m}{2}$. Here we set the reference point to $(-1,\ldots,-1)$ with $m$ dimensions.

\subsection{Runtime}

We now analyze the runtime of the \SMS with the aging strategy optimizing $\mojzj$. To facilitate the proof, we resort to the technique used in \cite{WiethegerD24} which uses a union bound to establish a tail bound of the runtime. Before the analysis, we first recall several definitions from \cite{WiethegerD24}. Let $m' = m/2$. The set $ K_{m,k}$ is the set of function values of individuals in which each block has either exactly $k$ 0-bits or $k$ 1-bits. Formally, $K_{m,k} = \left\{ (a_1, \ldots, a_{m'}) \mid a_i \in \left\{ k, \frac{n}{m'} - k \right\} \text{ for all } i \in [m'] \right\}.$ Similarly, the set $C_{m,k}$ is the set of function values of individuals in which each block consists either all bits being 1, all bits being 0, exactly $k$ 0-bits, or $k$ 1-bits. Formally, $C_{m,k} = \left\{ (a_1, \ldots, a_{m'}) \mid a_i \in \left\{ 0, k, \frac{n}{m'} - k, \frac{n}{m'} \right\} \text{ for all } i \in [m'] \right\}.$

We divide the optimization process into three phases. The first phase starts with the initialization of the algorithm and concludes when all points in $K_{m,k}$ are covered. The second phase begins thereafter and ends when all points in $C_{m,k}$ are covered. The final phase ends when the remaining Pareto front is covered. 

For the first phase and the final phase, we apply  Lemma~A1 and Lemma~A3 from \cite{WiethegerD24}, which provide the runtime analyses of the GSEMO on $\mojzj$. These results can be adapted to the \SMS with the aging strategy when the population size $\mu \geq \overline{M}+1+\tau$. The reason is that Section~5 of \cite{WiethegerD24} outlines three sufficient properties to prove such upper bounds. The survival guarantee provided by Lemma~\ref{lem:survival} satisfies the first one for not loosing already made progress. As for the remaining two properties, the probability of selecting an individual from the population for mutation is $1/\mu$ and standard bit-wise mutation is employed. Therefore, we obtain the following lemmas for the runtime of the first and final phases. 

\begin{lemma}
    Let $k\leq n'/2$. Consider the \SMS with the aging strategy and with population size $\mu \geq \overline{M}+1+\tau$ to optimize $\mojzj$. Let $T$ denote the number of iterations until the population covers $K_{m,k}$ and let
    \begin{align*}
        t = \left(\ln(2)\frac{m'}{\ln(n)}+2\right)e\mu(n-k)\ln(n-k).
    \end{align*}
    Then $T\leq t$ with probability at least $1-\frac{1}{n}$.
    \label{lem:many first phase}
\end{lemma}

\begin{lemma}
    Let $k\leq n'/2$. Consider the \SMS with the aging strategy and with population size $\mu \geq \overline{M}+1+\tau$ optimizing $\mojzj$ starting with a population that covers $C_{m,k}$. Let $T$ denote the number of iterations until the population covers the entire Pareto front and let
    \begin{align*}
    t = &2em'\mu \cdot \max \left\{ 2\left(\frac{n}{2m'}-k\right),\right. \\
        &\quad \left. 8 \ln (m') + 8m' \ln \left(\frac{n}{m'}-2k+3\right) + 8\ln(n) \right\}.
    \end{align*}
    Then $T\leq \lceil t \rceil$ with probability at least $1-\frac{1}{n}$.
    \label{lem:many final phase}
\end{lemma}

Now we consider the second phase that begins when all points in $K_{m,k}$ are covered and ends when all points in $C_{m,k}$ are covered. From the analysis of Lemma~\ref{lem:OJZJ second phase}, we know that the aging strategy helps to remain the inferior individuals, thereby increasing the probability of generating individuals with all 1-bits or all 0-bits. The following lemma shows that it continues to provide advantages in traversing fitness valleys for many-objective optimization. 
We first use the same approach to Lemma~\ref{lem:OJZJ second phase}, calculating the probability of changing a block with $n'-k$ 1-bits (or 0-bits) to the extreme block $1^{n'}$ (or $0^{n'}$) within at most $k\tau$ iterations. Then based on this, we calculate the tail bound for the time required to find the extreme solution $1^n$ which has $m'$ extreme blocks. A union bound is further used over the tail bounds to bound the time required to cover $C_{m,k}$. 

\begin{lemma}
    Let $k\leq n'/2$. Consider the \SMS with the aging strategy and with population size $\mu \geq \overline{M}+1+\tau$ optimizing $\mojzj$ starting with a population that covers $K_{m,k}$. Let $T$ denote the number of iterations until the population covers $C_{m,k}$ and let
    \begin{align*}
        t = &\left(1+\frac{\ln(4)m'+\ln(n)}{\ln(m')}\right)\frac{\ln(m')k\tau}{2}\\
        &\cdot\left(\frac{e^2 n}{k\left(1-\exp{\left(-\frac{\tau}{\mu}\right)}\right)}\right)^k .
    \end{align*}
    Then $T\leq t $ with probability at least $1-\frac{1}{n}$. Further, 
    \begin{align*}
        E[T]\leq &\left(1-\frac{1}{m'}\right)^{-1} \left(2+\frac{\ln(4)m'}{\ln(m')}\right)\frac{\ln(m')k\tau}{2}\\
        &\cdot\left(\frac{e^2 n}{k\left(1-\exp{\left(-\frac{\tau}{\mu}\right)}\right)}\right)^k.
    \end{align*}
    \label{lem:many second phase}
\end{lemma}

Combining Lemmas~\ref{lem:many first phase} to~\ref{lem:many second phase}, similarly using a simple restart argument to obtain the expected value for the first and final phase, we can easily derive the runtime of the \SMS with the aging strategy in the following theorem.

\begin{theorem}\label{thm:mojzj}
    Let $k\leq n'/2$. Consider the \SMS with the aging strategy that $\mu \geq \overline{M}+1+\tau$ to optimize $\mojzj$. Let T denote the number of iterations until the population covers the Pareto front and let 
    \begin{align*}
        t = &\left(1+\frac{\ln(4)m'+\ln(n)}{\ln(m')}\right)\frac{3\ln(m')k\tau}{2}\\
        &\cdot\left(\frac{e^2 n}{k\left(1-\exp{\left(-\frac{\tau}{\mu}\right)}\right)}\right)^k .
    \end{align*}
    Then $T\leq t$ with high probability. Further, 
    \begin{align*}
        E[T]\leq &\left(1-\frac{1}{m'}\right)^{-1} \left(2+\frac{\ln(4)m'}{\ln(m')}\right)\frac{3\ln(m')k\tau}{2}\\
        &\cdot\left(\frac{e^2 n}{k\left(1-\exp{\left(-\frac{\tau}{\mu}\right)}\right)}\right)^k .  
    \end{align*}
\end{theorem}

With $\tau = \Theta(\overline{M})$ and population size $\mu = \Theta(\overline{M}+1+\tau)$, the \SMS with the aging strategy requires $O(\overline{M}km(n/\Theta(k))^k)$ iterations in expectation to cover the full Pareto front of \mojzj. Compared to the runtime guarantee $O(\overline{M}mn^k)$ proved in \cite{WiethegerD24} for the original \SMS, the aging strategy achieves a speed-up by a factor of $\max\{1, \Theta(k)^{k-1}\}$. Together with Theorem~\ref{thm:ojzj} for bi-objective \ojzj, we see that our aging strategy achieves a scalable speed-up factor of $\max\{1,\Theta(k)^{k-1}\}$ for all number of objectives. It is a nice property that the stochastic population update strategy doesn't have (Note the speed-up reduces significantly for increasing the number of objectives, as mentioned before). Besides, a speed-up factor of $\max\{1, \Theta(k)^{k-1}\}$ for our aging strategy is also obtained compared to the stochastic population update.


\section{Experiments}
This section conducts experiments to intuitively see the efficiency of the aging strategy for two and more objectives. For comparison, we also include experiments for the original \SMS and the one with the stochastic population update.

For the bi-objective problem, we chose \ojzj as in Theorem~\ref{thm:ojzj}, and set the problem size $n \in \{10,15,20,25,30\}$ and gap size $k = 4$ to see whether the asymptotic results hold for small and medium problem sizes. We set the old enough age $\tau=\mu/2$ and the population size to $2(n-2k+4)$ for the best asymptotic runtime in Theorem~\ref{thm:ojzj}. 

\begin{figure}[htp]
    \centering
    \includegraphics[width=1.0\columnwidth]{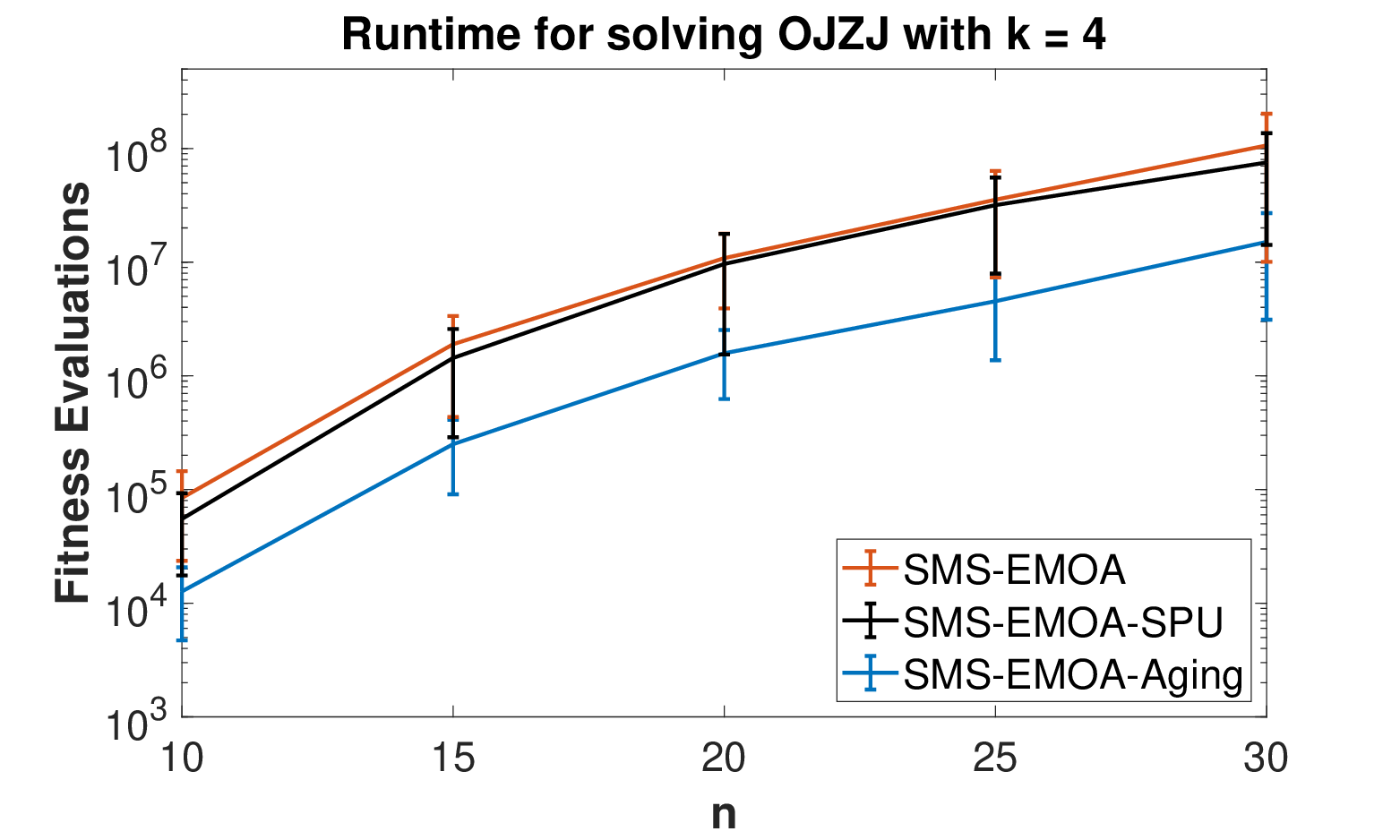}
    \caption{The mean (with standard deviations) number of fitness evaluations of the \SMS with different mechanisms for solving \ojzj with $k=4$ and $n\in\{10,15,20,25,30\}$ in 50 independent runs.}
    \label{fig:ojzjk4}
\end{figure}

Each algorithm was tested with 50 independent runs, and terminated when the full Pareto front was covered for the first time. Figure~\ref{fig:ojzjk4} illustrates the mean (with standard deviations) number of iterations for each algorithm to cover the full Pareto front for the first time. We easily see a speed-up of about $7$ of the aging strategy against the original \SMS and a speed-up of around $5$ against the stochastic population update. It shows that the superiority of the aging strategy already appears for small problem sizes and small gap size. 

\begin{figure}[htp]
    \centering
    \includegraphics[width=1.0\columnwidth]{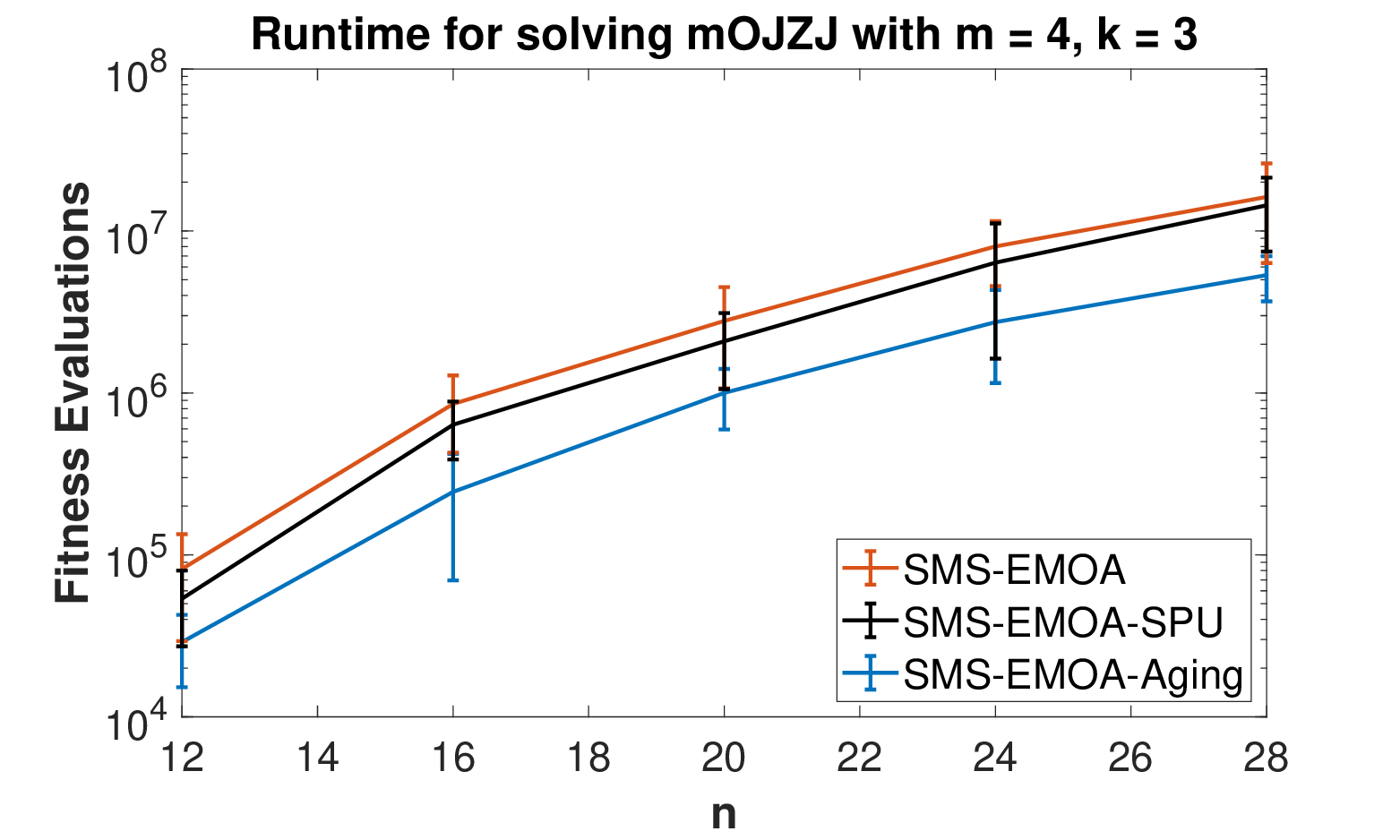}
    \caption{The mean (with standard deviations) number of fitness evaluations of the \SMS with different mechanisms for solving \mojzj with $m=4,k=3$ and $n\in\{12,16,20,24,28\}$ in 20 independent runs.}
    \label{fig:mojzj}
\end{figure}

For many-objective optimization, we chose \mojzj as in Theorem~\ref{thm:mojzj}, and set the number of objectives $m = 4$. We do not set larger numbers of objectives, since (i) experiments with $m=4$ finished in a reasonable time and (ii) with $m=4$ we intend to see whether our aging strategy can result in a good speed-up even for small number of objectives. We set the problem size $n\in\{12,16,20,24,28\}$ and gap size $k=3$ to see whether the asymptotic results hold for small and medium problem sizes. The old enough age $\tau$ was still set to $\mu/2$ and the population size was set to $2\left(\left(n'+1\right)^{m/2}+1\right)$ for the best asymptotic runtime in Theorem~\ref{thm:mojzj}. Each algorithm was tested with $20$ independent runs, and terminated when the full Pareto front was covered for the first time. From Figure~\ref{fig:mojzj}, we easily see the clear superiority of the \SMS with the aging strategy against the original one and the one with the stochastic population update. Note that the theoretical speed-up factor against the other two is $\Theta(k)^{k-1}$ from Theorem~\ref{thm:mojzj}. 
For this setting, $k^{k-1}=3^2=9$. 
From the plot, we know such speed-up factor is around $3$ compared to the original \SMS and is around $2$ compared to the one with the stochastic population update, which means that the theoretical speed-up is already witnessed for small problem sizes, small gap size, and small number of objectives. 

\section{Conclusion}
In this paper, to tackle the quest for a scalably efficient non-elitist selection strategy for MOEAs, we proposed a simple aging strategy, essentially giving each new individual $\tau$ iterations to develop before being prone to removal, and added it to the \SMS. A speed-up of $\max\{1, \Theta(k)^{k-1}\}$ compared to the original \SMS (and also the one with the stochastic population update) is proven for the $m$-objective \jump benchmark for all even $m$. This is the first non-elitist strategy for which a speed-up not vanishing for larger numbers of objectives is proven. 

Also for the bi-objective \jump benchmark, our speed-up of $\max\{1, \Theta(k)^{k-1}\}$ compares favorably with the previously shown speed-up of $\max\{1,2^k / n\}$, and in particular means that a positive speed-up is seen for small values of $k$, which are most relevant (only constant $k$ give a polynomial runtime). Our experiments support the above findings.

From our understanding of the mathematical proofs, we conjecture that our simple non-elitist aging strategy can be effective in other MOEAs as well, and list this as our most interesting future work. 

\section*{Acknowledgments}
This work was supported by National Natural Science Foundation of China (Grant No. 62306086, 62350710797), Science, Technology and Innovation Commission of Shenzhen Municipality (Grant No. GXWD20220818191018001), and Guangdong Basic and Applied Basic Research Foundation (Grant No. 2025A1515011936). This research benefited from the support of the FMJH Program PGMO.


\end{document}